\documentclass[11pt]{article}
\usepackage{coling2018}
\usepackage{times}
\usepackage{url}
\usepackage{latexsym}

\usepackage{bbm}
\usepackage{bm}
\usepackage{url}
\usepackage{amsmath}
\usepackage{amssymb}
\usepackage{graphicx} 
\usepackage{color}
\usepackage{subcaption}

\newcommand{\vect}[1]{\bm{#1}}
\newcommand{\mat}[1]{\bf{#1}}
\newcommand{\RR}{\mathbb{R}}

\title{Enhancing Sentence Embedding with Generalized Pooling}

\author{
Qian Chen \\
University of Science and \\ 
Technology of China\\
\tt{cq1231@mail.ustc.edu.cn} \\\And
Zhen-Hua Ling \\
University of Science and \\
Technology of China\\
\tt{zhling@ustc.edu.cn} \\\AND
Xiaodan Zhu \\
ECE, Queen's University \\
\texttt{xiaodan.zhu@queensu.ca} \\
}
\date{}

\begin{document}
\maketitle
\begin{abstract}
Pooling is an essential component of a wide variety of sentence representation and embedding models. This paper explores generalized pooling methods to enhance sentence embedding. We propose vector-based multi-head attention that includes the widely used max pooling, mean pooling, and scalar self-attention as special cases. The model benefits from properly designed penalization terms to reduce redundancy in multi-head attention. We evaluate the proposed model on three different tasks: natural language inference (NLI), author profiling, and sentiment classification. The experiments show that the proposed model achieves significant improvement over strong sentence-encoding-based methods, resulting in state-of-the-art performances on four datasets. The proposed approach can be easily implemented for more problems than we discuss in this paper.

\end{abstract}

\section{Introduction}
%
% The following footnote without marker is needed for the camera-ready
% version of the paper.
% Comment out the instructions (first text) and uncomment the 8 lines
% under "final paper" for your variant of English.
% 
\blfootnote{
    %
    % for review submission
    %
%     \hspace{-0.65cm}  % space normally used by the marker
%     Place licence statement here for the camera-ready version. See
%     Section~\ref{licence} of the instructions for preparing a
%     manuscript.
    %
    % % final paper: en-uk version 
    %
    % \hspace{-0.65cm}  % space normally used by the marker
    % This work is licensed under a Creative Commons 
    % Attribution 4.0 International Licence.
    % Licence details:
    % \url{http://creativecommons.org/licenses/by/4.0/}.
    % 
    % % final paper: en-us version 
    %
    \hspace{-0.65cm}  % space normally used by the marker
    This work is licensed under a Creative Commons 
    Attribution 4.0 International License.
    License details:
    \url{http://creativecommons.org/licenses/by/4.0/}.
}

Distributed representation learned with neural networks has shown to be effective in modeling natural language at different granularities. Learning representation for words~\cite{DBLP:conf/nips/BengioDV00,DBLP:journals/corr/abs-1301-3781,DBLP:conf/emnlp/PenningtonSM14}, for example, has achieved notable success. Much remains to be done to model larger spans of text such as sentences or documents. 
The approaches to computing sentence embedding generally fall into two categories. The first consists of learning sentence embedding with unsupervised learning, e.g., auto-encoder-based models~\cite{DBLP:conf/emnlp/SocherPHNM11}, Paragraph Vector~\cite{DBLP:conf/icml/LeM14}, SkipThought vectors~\cite{DBLP:conf/nips/KirosZSZUTF15}, FastSent~\cite{DBLP:conf/naacl/HillCK16}, among others. 
The second category consists of models trained with supervised learning, such as convolution neural networks (CNN)~\cite{DBLP:conf/emnlp/Kim14,DBLP:conf/acl/KalchbrennerGB14}, recurrent neural networks (RNN)~\cite{DBLP:conf/emnlp/ConneauKSBB17,DBLP:conf/emnlp/BowmanAPM15}, and tree-structure recursive networks~\cite{D13-1170,DBLP:conf/icml/ZhuSG15,DBLP:conf/acl/TaiSM15}, just to name a few. 

Pooling is an essential component of a wide variety of sentence representation and embedding models. For example, in recurrent-neural-network-based models, pooling is often used to aggregate hidden states at different time steps (i.e., words in a sentence) to obtain sentence embedding. Convolutional neural networks (CNN) also often uses max or mean pooling to obtain a fixed-size sentence embedding. 

In this paper we explore generalized pooling methods to enhance sentence embedding. Specifically, by extending scalar self-attention models such as those proposed in~\newcite{DBLP:journals/corr/LinFSYXZB17}, we propose vector-based multi-head attention, which includes the widely used max pooling, mean pooling, and scalar self-attention itself as special cases. On one hand, the proposed method allows for extracting different aspects of the sentence into multiple vector representations through the multi-head mechanism. On the other, it allows the models to focus on one of many possible interpretations of the words encoded in the context vector through the vector-based attention mechanism. In the proposed model we design penalization terms to reduce redundancy in multi-head attention. 

We evaluate the proposed model on three different tasks: natural language inference, author profiling, and sentiment classification. The experiments show that the proposed model achieves significant improvement over strong sentence-encoding-based methods, resulting in state-of-the-art performances on four datasets. The proposed approach can be easily implemented for more problems than we discuss in this paper.

\section{Related Work}

There exist in the literature much previous work for sentence embedding with supervised learning, which mostly use RNN and CNN as building blocks. For example,~\newcite{DBLP:conf/emnlp/BowmanAPM15} used BiLSTMs as sentence embedding for natural language inference task.~\newcite{DBLP:conf/emnlp/Kim14} used CNN with max pooling for sentence classification. More complicated neural networks were also proposed for sentence embedding. For example,~\newcite{D13-1170} introduced Recursive Neural Tensor Network (RNTN) over parse trees to compute sentence embedding for sentiment analysis.~\newcite{DBLP:conf/icml/ZhuSG15} and~\newcite{DBLP:conf/acl/TaiSM15} proposed tree-LSTM.~\newcite{DBLP:conf/eacl/YuM17a} proposed a memory augmented neural networks, called Neural Semantic Encoder (NSE), as sentence embedding for natural language understanding tasks.

Some recent research began to explore inner/self-sentence attention mechanism for sentence embedding, which can be classified into two categories: self-attention network and self-attention pooling.~\newcite{DBLP:conf/emnlp/0001DL16} proposed an intra-sentence level attention mechanism on the base of LSTM, called LSTMN. For each step in LSTMN, it calculated the attention between a certain word and its previous words.~\newcite{DBLP:conf/nips/VaswaniSPUJGKP17} proposed a self-attention network for the neural machine translation task. The self-attention network uses multi-head scaled dot-product attention to represent each word by weighted summation of all word in the sentence.~\newcite{DBLP:journals/corr/abs-1709-04696} proposed DiSAN, which is composed of a directional self-attention with temporal order encoded.~\newcite{DBLP:journals/corr/abs-1801-10296} proposed reinforced self-attention network (ReSAN), which integrate
both soft and hard attention into one context fusion with reinforced learning. 

Self-attention pooling has also been studied in previous work.~\newcite{DBLP:journals/corr/LiuSLW16} proposed inner-sentence attention based pooling methods for sentence embedding. They calculate scalar attention between the LSTM states and the mean pooling using multi-layer perceptron (MLP) to obtain the vector representation for a sentence.~\newcite{DBLP:journals/corr/LinFSYXZB17} proposed a scalar structure/multi-head self-attention method for sentence embedding. The multi-head self-attention is calculated by a MLP with only LSTM states as input. There are two main differences from our proposed method; i.e., (1) they used scalar attention instead of vectorial attention, (2) we propose different penalization terms which is suitable for vector-based multi-head self-attention, while their penalization term on attention matrix is only designed for scalar multi-head self-attention.~\newcite{choi2018fine} proposed a fine-grained attention mechanism for neural machine translation, which also extend scalar attention to vectorial attention.~\newcite{DBLP:journals/corr/abs-1709-04696} proposes multi-dimensional/vectorial self-attention pooling on the top of self-attention network instead of BiLSTM. However, both of them didn't consider multi-head self-attention.

\section{The Model}
In this section we describe the proposed models that enhance sentence embedding with generalized pooling approaches. The pooling layer is built on a state-of-the-art sequence encoder layer. Below, we first discuss the sequence encoder, which, when enhanced with the proposed generalized pooling, achieves state-of-the-art performance on three different tasks on four datasets.  

\subsection{Sequence Encoder}
%\subsection{Word Embedding}

The sequence encoder in our model takes into $T$ word tokens of a sentence ${\vect S} = (w_1, w_2, \dots, w_T)$. Each word $w_i$ is from the vocabulary ${\vect V}$. For each word we concatenate pre-trained word embedding and embedding learned from characters. The character composition model feeds all characters of the word into a convolution neural network (CNN) with max pooling~\cite{DBLP:conf/emnlp/Kim14}. The detailed experiment setup will be discussed in Section \ref{sec:setup}. The sentence ${\vect S}$ is represented as a word embedding sequence: ${\vect X} = ({\vect e}_1, {\vect e}_2, \dots, {\vect e}_T) \in \RR^{T \times d_e}$, where $d_e$ is the dimension of word embedding which concatenates embedding obtained from character composition and pretrained word embedding.

To represent words and their context in sentences, the sentences are fed into stacked bidirectional LSTMs (BiLSTMs). Shortcut connections are applied, which concatenate word embeddings and input hidden states at each layer in the stacked BiLSTM except for the first (bottom) layer. The formulae are as follows:

\begin{align}
\overrightarrow{\vect h}^l_t &= \text{LSTM}([{\vect e}_t;\overrightarrow{\vect h}^{l-1}_t], \overrightarrow{\vect h}_{t-1}^l) \,,\\
\overleftarrow{\vect h}^l_t &= \text{LSTM}([{\vect e}_t;\overleftarrow{\vect h}^{l-1}_t], \overleftarrow{\vect h}_{t+1}^l) \,,\\
{\vect h}_t^l &= [\overrightarrow{\vect h}_t^l;\overleftarrow{\vect h}_t^l]\,.
\end{align}

\noindent where hidden states ${\vect h}_t^l$ in layer $l$ concatenate two directional hidden states of LSTM at time $t$.
Then the sequence is represented as the hidden states in the top layer $L$: ${\mat H}^{L} = ({\vect h}_1^L, {\vect h}_2^L,\dots, {\vect h}_T^L) \in \RR^{T \times 2d}$. For simplicity, we ignore the superscript $L$ in the remainder of the paper.

\subsection{Generalized Pooling}
\subsubsection{Vector-based Multi-head Attention} 
To transform a variable length sentence into a fixed size vector representation, we propose a generalized pooling method. We achieve that by using a weighted summation of the $T$ LSTM hidden vectors, and the weights are vectors rather than scalars, which can control every element in all hidden vectors: 
\begin{align}
{\mat A} = \text{softmax}({\mat W}_2\text{ReLU}({\mat W}_1 {\mat H}^\mathrm{T} + {\vect b_1} ) + {\vect b_2})^\mathrm{T} \,,
\end{align}

\noindent where ${\mat W}_1 \in \RR^{d_a \times 2d}$ and ${\mat W}_2 \in \RR^{2d \times d_a}$ are weight matrices; ${\vect b_1} \in \RR^{d_a}$ and ${\vect b_2} \in \RR^{2d}$ are bias, where $d_a$ is the dimension of attention network and $d$ is the dimension of LSTMs. ${\mat H} \in \RR^{T \times 2d}$ and ${\mat A} \in \RR^{T \times 2d}$ are the hidden vectors at the top layer and weight matrices, respectively. The softmax ensures that $({\mat A}_1, {\mat A}_2, \dots, {\mat A}_T)$ are non-negative and sum up to 1 for every element in vectors. Then we sum up the LSTM hidden states ${\mat H}$ according to the weight vectors provided by ${\mat A}$ to get a vector representation ${\vect v}$ of the input sentence.

However, the vector representation usually focuses on a specific component of the sentence, like a special set of related words or phrases. We extend pooling method to a multi-head way: 

\begin{align}
\label{equ:att}
{\mat A}^i &= \text{softmax}({\mat W}^i_2\text{ReLU}({\mat W}^i_1 {\mat H}^\mathrm{T} + {\vect b^i_1} ) + {\vect b^i_2})^\mathrm{T} \,, \forall i \in {1,\dots, I} \,, \\
{\vect v}^i &= \sum_{t=1}^{T}{\vect a}^i_t \odot {\vect h^i_t}  \,, \forall i \in {1,\dots, I} \,,
\end{align}

\noindent where ${\vect a}^i_t$ indicates the vectorial attention from ${\mat A}^i$ for the $t$-th token in $i$-th head and $\odot$ is the element-wise product (also called the Hadamard product). Thus the final representation is a concatenated vector ${\vect v} = [{\vect v}^1;{\vect v}^2;\dots;{\vect v}^I]$, where each ${\vect v}^i$ captures different aspects of the sentence. For example, some heads of vectors may represent the predicate of sentence and other heads of vectors represent argument of the sentence, which enhances representation of sentences obtained in single-head attention.

\subsubsection{Penalization Terms}

To reduce the redundancy of multi-head attention, we design penalization terms for vector-based multi-head attention in order to encourage the diversity of summation weight across different heads of attention. We propose three types of penalization terms.

\paragraph{Penalization Term on Parameter Matrices}
The first penalization term is applied to parameter matrix ${\mat W}^i_1$ in Equation~\ref{equ:att}, as shown in the following formula:

\begin{align}
P &= \mu \sum_{i=1}^{I}\sum_{j=i+1}^{I}\max(\lambda -\lVert {\mat W}^i_1 - {\mat W}^j_1 \rVert_\mathrm{F}^2, 0) \,.
\end{align}

Intuitively, we encourage different heads to have different parameters. We maximum the \textit{Frobenius} norm of the differences between two parameter matrices, resulting in encouraging the diversity of different heads.
It has no further bonus when the \textit{Frobenius} norm of the difference of two matrices exceeds the a threshold $\lambda$.
Similar to adding an L2 regularization term on neural networks, the penalization term $P$ will be added to the original loss with a weight of $\mu$. Hyper-parameters $\lambda$ and $\mu$ need to be tuned on a development set. We can also add constrains on ${\mat W}^i_2$ in a similar way, but we did not observe further improvement in our experiments. 

\paragraph{Penalization Term on Attention Matrices}
The second penalization term is added on attention matrices. Instead of using $\lVert{\mat A} {\mat A}^\mathrm{T} - {\mat I}\rVert_\mathrm{F}^2$ to encourage the diversity for scalar attention matrix as in~\newcite{DBLP:journals/corr/LinFSYXZB17}, we propose the following formula to encourage the diversity for vectorial attention matrices. The penalization term on attention matrices is 
\begin{align}
P = \mu \sum_{i=1}^{I}\sum_{j=i+1}^{I}\max(\lambda -\lVert {\mat A}^i - {\mat A}^j \rVert_\mathrm{F}^2, 0) \,,
\end{align}

\noindent where $\lambda$ and $\mu$ are hyper-parameters which need to be tuned based on a development set. Intuitively, we try to encourage the  diversity of any two different ${\mat A}^i$ under the threshold $\lambda$. 

\paragraph{Penalization Term on Sentence Embeddings}
In addition, we propose to add a penalization term on multi-head sentence embedding ${\vect v}^i$ directly as follows:

\begin{align}
P = \mu \sum_{i=1}^{I}\sum_{j=i+1}^{I}\max(\lambda - \lVert {\vect v}^i - {\vect v}^j \rVert_2^2, 0) \,,
\end{align}
\noindent where $\lambda$ and $\mu$ are hyper-parameters. Here we try to maximize the $l^2$-norm of any two different heads of sentence embeddings under the threshold $\lambda$.

\subsection{Top-layer Classifiers}

The output of pooling is fed to a top-layer classifier to solve different problems. In this paper we evaluate our sentence embedding models on three different tasks: natural language
inference (NLI), author profiling, and sentiment classification, on four datasets. The evaluation covers two typical types of problems. The author profiling and sentiment tasks classify individual sentences into different categories and the two NLI tasks classify sentence pairs. 

For the NLI tasks, to enhance the relationship between sentence pairs, we concatenate the embeddings of two sentences with their absolute difference and element-wise product~\cite{DBLP:conf/acl/MouMLX0YJ16} as the input to the multilayer perceptron (MLP) classifier:

\begin{equation}
{\vect v} = [{\vect v}_a;{\vect v}_b;\lvert{\vect v}_a - {\vect v}_b\rvert; {\vect v}_a \odot {\vect v}_b] \,,
\end{equation}

\noindent where $\odot$ is the element-wise product. The MLP has two hidden layers with \textit{ReLU} activation with shortcut connections and a \textit{softmax} output layer. The entire model is trained end-to-end through minimizing the cross-entropy loss. Note that for the two classification tasks on individual sentences (i.e., the author profiling and sentiment classification task), we use the same MLP classifiers described above for sentence pair classification. But instead of concatenating two sentences, we directly feed a sentence embedding into MLP.

\section{Experimental Setup}
\label{sec:setup}

\subsection{Data}
\paragraph{SNLI} The SNLI~\cite{DBLP:conf/emnlp/BowmanAPM15} is a large dataset for natural language inference. The task detects three relationships between a premise and a hypothesis sentence: the premise entails the hypothesis (\textit{entailment}), they contradict each other (\textit{contradiction}), or they have a neutral relation (\textit{neutral}). We use the same data split as in~\newcite{DBLP:conf/emnlp/BowmanAPM15}, i.e., 549.367 samples for training, 9,842 samples for development and 9,824 samples for testing. 

\paragraph{MultiNLI} MultiNLI~\cite{DBLP:journals/corr/WilliamsNB17} is another natural language inference dataset. The data are collected from a broader range of genres such as fiction, letters, telephone speech, and 9/11 reports. Half of these 10 genres are used in training while the rest are not, resulting in-domain and cross-domain development and test sets used to test NLI systems. We use the same data split as in~\newcite{DBLP:journals/corr/WilliamsNB17}, i.e., 392,702 samples for training, 9,815/9,832 samples for in-domain/cross-domain development, and 9,796/9,847 samples for in-domain/cross-domain testing. Note that, we do not use SNLI as an additional training/development set in our experiments.

\paragraph{Age Dataset} To compare our models with that of~\newcite{DBLP:journals/corr/LinFSYXZB17}, we use the same Age dataset in our experiment here, which is an Author Profiling dataset. The dataset are extracted from the Author Profiling dataset\footnote{http://pan.webis.de/clef16/pan16-web/author-profiling.html}, which consists of tweets from English Twitter. The task is to predict the age range of authors of input tweets. The age range are split into 5 classes: 18-24, 25-34, 35-49, 50-64, 65+. We use the same data split as in~\newcite{DBLP:journals/corr/LinFSYXZB17}, i.e., 68,485 samples for training, 4,000 for development, and 4,000 for testing. 

\paragraph{Yelp Dataset} The Yelp dataset\footnote{https://www.yelp.com/dataset/challenge} is a sentiment analysis task, which takes reviews as input and predicts the level of sentiment in terms of the number of stars, from 1 to 5 stars, where 5-star means the most positive. We use the same data split as in~\newcite{DBLP:journals/corr/LinFSYXZB17}, i.e., 500,000 samples for training, 2,000 for development, and 2,000 for testing.

\subsection{Training Details} 
We implement our algorithm with Theano~\cite{2016arXiv160502688short} framework.
We use the development set (in-domain development set for MultiNLI) to select models for testing. To help replicate our results, we publish our code\footnote{https://github.com/lukecq1231/generalized-pooling}, which is developed from our codebase for multiple tasks~\cite{DBLP:conf/acl/ChenZLIW18,DBLP:conf/acl/ChenZLWJI17,DBLP:conf/ijcai/ChenZLWJ16,Zhang:qa:2017}. Specifically, we use Adam~\cite{DBLP:journals/corr/KingmaB14} for optimization. The initial learning rate is 4e-4 for SNLI and MultiNLI, 2e-3 for Age dataset, 1e-3 for Yelp dataset. For SNLI and MultiNLI dataset, stacked BiLSTMs have 3 layers. For Age and Yelp dataset, stacked BiLSTMs have 1 layer. The hidden states of BiLSTMs for each direction and MLP are 300 dimension, except for SNLI whose dimensions are 600. We clip the norm of gradients to make it smaller than 10 for SNLI and MultiNLI, and 0.5 for Age and Yelp dataset. The character embedding has 15 dimensions, and 1D-CNN filters lengths are 1, 3 and 5, respectively. Each filter has 100 feature maps, resulting in 300 dimensions for character-composition embedding. We initialize word-level embedding with pre-trained \textit{GloVe-840B-300D} embeddings~\cite{DBLP:conf/emnlp/PenningtonSM14} and initialize out-of-vocabulary words randomly with a Gaussian distribution. The word-level embedding is fixed during training for SNLI and MultiNLI dataset, but updated during training for Age and Yelp dataset, which is determined by the performance on development sets. 
The mini-batch size is 128 for SNLI and 32 for the rest. We use 5 heads generalized pooling for all tasks. And $d_a$ is 600 for SNLI and 300 for the other datasets. For the penalization term, we choose $\lambda = 1$; the penalization weight $\mu$ is selected from [1,1e-1,1e-2,1e-3,1e-4] based on performances on the development sets.

\section{Experimental Results}
\subsection{Overall Performance}
For the NLI tasks, there are many ways to add cross-sentence~\cite{DBLP:journals/corr/RocktaschelGHKB15,DBLP:conf/emnlp/ParikhT0U16,DBLP:conf/acl/ChenZLWJI17} level attention. To ensure the comparison is fair, we only compare methods that use sentence-encoding-based models; i.e., cross-sentence attention is not allowed. Note that this follows the setup in the RepEval-2017 Shared Task. Table~\ref{tab:snli} shows the results of different models for NLI, consisting of results of previous work on sentence-encoding-based models, plus the performance of our baselines and that of the model proposed in this paper. We have three additional baseline models: the first uses max pooling on top of BiLSTM, which achieves an accuracy of 85.3\%; the second uses mean pooling on top of BiLSTM, which achieves an accuracy of 85.0\%; the third uses last pooling, i.e., concatenating the last hidden states of forward and backward LSTMs, which achieves an accuracy of 84.9\%. Instead of using heuristic pooling methods, the proposed sentence-encoding-based model with generalized pooling achieves a new state-of-the-art accuracy of 86.6\% on the SNLI dataset; the improvement over the baseline with max pooling is statistically significant under the one-tailed paired t-test at the 99.999\% significance level. The previous state-of-the-art model ReSAN~\cite{DBLP:journals/corr/abs-1801-10296} used a hybrid of hard and soft attention model with reinforced learning achieved an accuracy of 86.3\%.  

\begin{table}[t!]
\renewcommand{\arraystretch}{0.9}
\begin{center}
\scalebox{1}{
\begin{tabular}{|l|r|}
\hline \bf Model & \bf Test \\ \hline
100D LSTM~\cite{DBLP:conf/emnlp/BowmanAPM15} & 77.6 \\
300D LSTM~\cite{DBLP:conf/acl/BowmanGRGMP16} & 80.6 \\
1024D GRU~\cite{DBLP:journals/corr/VendrovKFU15} & 81.4 \\
300D Tree CNN~\cite{DBLP:conf/acl/MouMLX0YJ16} & 82.1 \\
600D SPINN-PI~\cite{DBLP:conf/acl/BowmanGRGMP16} & 83.3 \\
600D BiLSTM~\cite{DBLP:journals/corr/LiuSLW16} & 83.3 \\
300D NTI-SLSTM-LSTM~\cite{DBLP:conf/eacl/YuM17} & 83.4 \\
600D BiLSTM intra-attention~\cite{DBLP:journals/corr/LiuSLW16} & 84.2 \\
600D BiLSTM self-attention~\cite{DBLP:journals/corr/LinFSYXZB17} & 84.4 \\
4096D BiLSTM max pooling~\cite{DBLP:conf/emnlp/ConneauKSBB17} & 84.5 \\
300D NSE~\cite{DBLP:conf/eacl/YuM17a} & 84.6 \\
600D BiLSTM gated-pooling~\cite{DBLP:conf/repeval/ChenZLWJI17} & 85.5 \\
300D DiSAN~\cite{DBLP:journals/corr/abs-1709-04696} & 85.6 \\
300D Gumbel TreeLSTM~\cite{DBLP:journals/corr/ChoiYL17} & 85.6 \\ 
600D Residual stacked BiLSTM~\cite{DBLP:conf/repeval/NieB17} & 85.7 \\
300D CAFE~\cite{DBLP:journals/corr/abs-1801-00102} & 85.9 \\
600D Gumbel TreeLSTM~\cite{DBLP:journals/corr/ChoiYL17} & 86.0 \\ 
1200D Residual stacked BiLSTM~\cite{DBLP:conf/repeval/NieB17} & 86.0 \\
300D ReSAN~\cite{DBLP:journals/corr/abs-1801-10296} & \textbf{86.3} \\
\hline
1200D BiLSTM max pooling & \underline{85.3} \\
1200D BiLSTM mean pooling & 85.0 \\
1200D BiLSTM last pooling & 84.9 \\
1200D BiLSTM \textbf{generalized pooling} & \textbf{86.6} \\
\hline
\end{tabular}
}
\end{center}
\caption{Accuracies of the models on the SNLI dataset. }
\label{tab:snli}
\end{table}

Table~\ref{tab:multinli} shows the results of different models on the MultiNLI dataset. The first group is the results of previous sentence-encoding-based models. The proposed model with generalized pooling achieves an accuracy of 73.8\% on the in-domain test set and 74.0\% on the cross-domain test set; both improve over the baselines using max pooling, mean pooling and last pooling. In addition, the results on cross-domain test set yield a new state of the art at an accuracy of 74.0\%, which is better than 73.6\% of shortcut-stacked BiLSTM~\cite{DBLP:conf/repeval/NieB17}.

\begin{table}[t!]
\centering
\begin{tabular}{|l|c|c|}
\hline
\textbf{Model}     & \textbf{In} & \textbf{Cross}   \\
\hline
CBOW~\cite{DBLP:journals/corr/WilliamsNB17} & 64.8 & 64.5 \\
BiLSTM~\cite{DBLP:journals/corr/WilliamsNB17} &  66.9 & 66.9\\
BiLSTM gated-pooling~\cite{DBLP:conf/repeval/ChenZLWJI17} & 73.5 & \textbf{73.6} \\
Shortcut stacked BiLSTM~\cite{DBLP:conf/repeval/NieB17} & \textbf{74.6} & \textbf{73.6} \\
\hline
BiLSTM max pooling & \underline{73.6} & \underline{73.0}\\
BiLSTM mean pooling & 71.5 & 71.6 \\
BiLSTM last pooling & 71.6 & 71.9\\
BiLSTM \textbf{generalized pooling} & \textbf{73.8} &\textbf{74.0} \\
\hline
\end{tabular}
\caption{Accuracies of the models on the MultiNLI dataset.}
\label{tab:multinli}
\end{table}
\begin{table}[t!]
\centering
\begin{tabular}{|l|c|c|}
\hline
\textbf{Model}     & \textbf{Yelp} & \textbf{Age} \\
\hline
BiLSTM max pooling~\cite{DBLP:journals/corr/LinFSYXZB17} & 61.99 & 77.30 \\
CNN max pooling~\cite{DBLP:journals/corr/LinFSYXZB17} & 62.05 & 78.15 \\
BiLSTM self-attention~\cite{DBLP:journals/corr/LinFSYXZB17} & \textbf{64.21} & \textbf{80.45} \\
\hline
BiLSTM max pooling & 65.00 & \underline{82.30}\\
BiLSTM mean pooling & \underline{65.30}  & 81.78 \\
BiLSTM last pooling & 64.95 & 81.10\\
BiLSTM \textbf{generalized pooling} & \textbf{66.55}  & \textbf{82.63}\\
\hline
\end{tabular}
\caption{Accuracies of the models on the Yelp and Age dataset. }
\label{tab:yelp_age}
\end{table}

Table~\ref{tab:yelp_age} shows the results of different models for the Yelp and the Age dataset. The BiLSTM with self-attention proposed by~\newcite{DBLP:journals/corr/LinFSYXZB17} achieves better result than CNN and BiLSTM with max pooling. One of our baseline models using max pooling on BiLSTM achieves accuracies of 65.00\% and 82.30\% on the Yelp and the Age dataset respectively, which is already better than the self-attention model proposed by~\newcite{DBLP:journals/corr/LinFSYXZB17}. We also show that the results of baseline with mean pooling and last pooling, in which mean pooling achieves the best result on the Yelp dataset among three baseline models and max pooling achieves the best on the Age dataset among three baselines. Our proposed generalized pooling method obtains further improvement on these already strong baselines, achieving 66.55\% on the Yelp dataset and 82.63\% on the Age dataset (statistically significant $p < 0.00001$ against best baselines), which are also new state of the art performances on these two datasets. 

\subsection{Detailed Analysis}

\begin{figure}[!htb]
	\centering
	\includegraphics[width=0.5\linewidth]{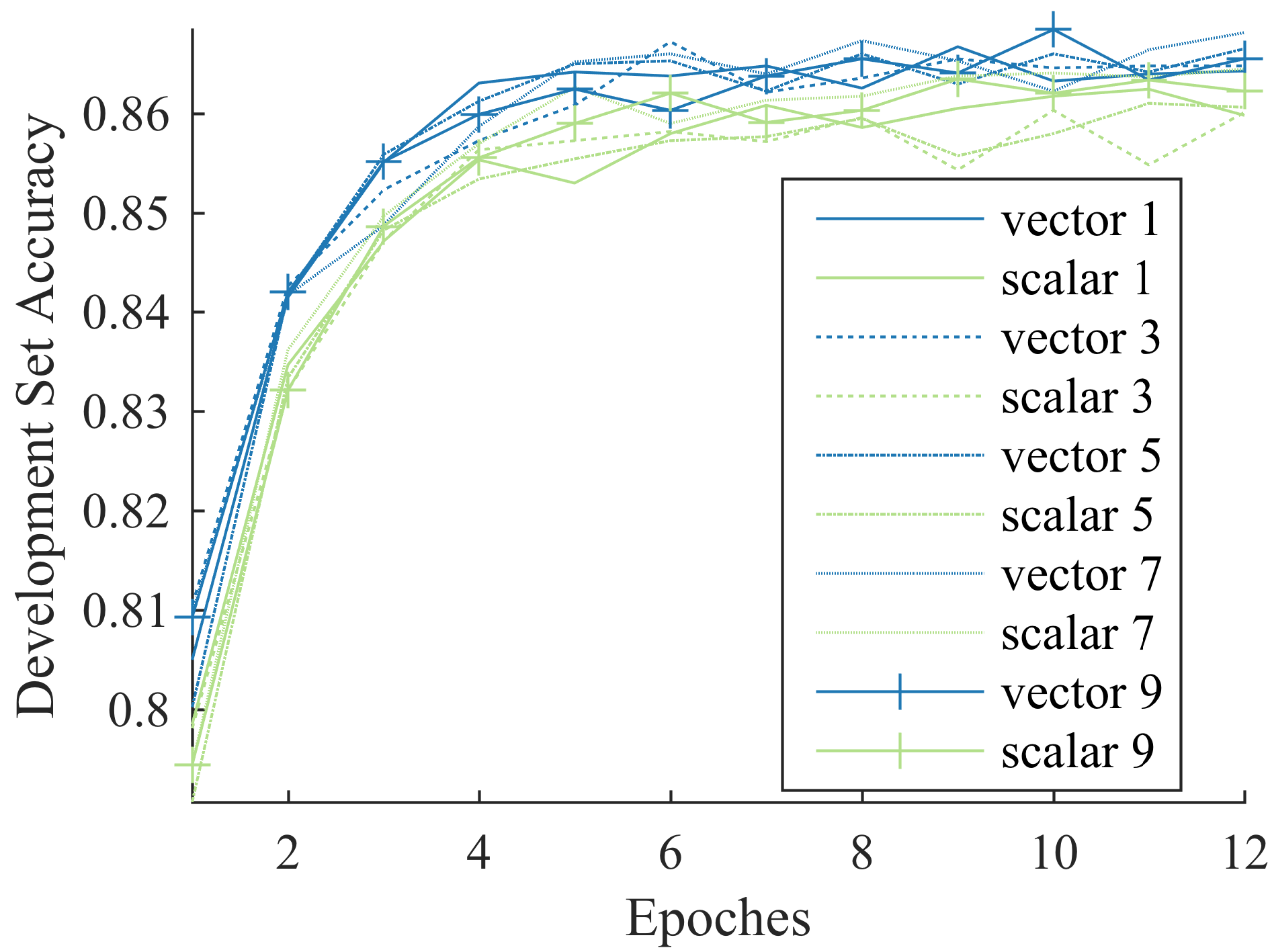}
	\caption{The effect of the number of heads and vectors/scalars in sentence embedding. The vertical axis indicates the development-set accuracy and the horizontal axis indicates training epochs. Numbers in the legend are the number of heads.}
	\label{fig:curve}
\end{figure}

\paragraph{Effect of Multiple Vectors/Scalars}
To compare the difference between vector-based attention and scalar attention, we draw the learning curves of different models using different heads on the SNLI development dataset without penalization terms as in Figure~\ref{fig:curve}. The green lines indicate scalar self-attention pooling added on top of the BiLSTMs, same as in~\newcite{DBLP:journals/corr/LinFSYXZB17}, and the blue lines indicate vector-based attention used in our generalized pooling methods. It is obvious that the vector-based attention achieves improvement over scalar attention. Different line styles are used to indicate self-attention using different numbers of multi-head, ranging from 1, 3, 5, 7 to 9. For vector-based attention, the 9-head model achieves the best accuracy of 86.8\% on the development set. For scalar attention, the 7-head model achieves the best accuracy of 86.4\% on the development set.

\paragraph{Effect of Penalization Terms}
To analyze the effect of penalization terms, we show the results with/without penalization terms on the four datasets in Table~\ref{tab:penalization}. Without using any penalization terms, the proposed generalized pooling achieves an accuracy of 86.4\% on the SNLI dataset, which is already slightly better than previous models (compared to accuracy 86.3\% in~\newcite{DBLP:journals/corr/abs-1801-10296}). When we use penalization on parameter matrices, the proposed model achieves a further improvement with an accuracy of 86.6\%. In addition, we also observe a significant improvement on MultiNLI, Yelp and Age dataset after using the penalization terms. For the MultiNLI dataset, the proposed model with penalization on parameter matrices achieves an accuracy of 73.8\% and 74.0\% on the in-domain and the cross-domain test set, respectively, which outperform the accuracy of 73.7\% and 73.4\% of the model without penalization, respectively. For the Yelp dataset, the proposed model with penalization on parameter matrices achieves the best results among the three penalization methods, which also improve the accuracy of 65.25\% to 66.55\% compared to the models without penalization. For the Age dataset, the proposed model with penalization on attention matrices achieves the best accuracy of 82.63\%, compared to the 82.18\% accuracy of the model without penalization. In general, the penalization on parameter matrices achieves the most effective improvement among most of these tasks, except for the Age dataset.

\begin{figure}[!htb]

\begin{subfigure}{1\textwidth}
  \centering
  \includegraphics[width=1\linewidth]{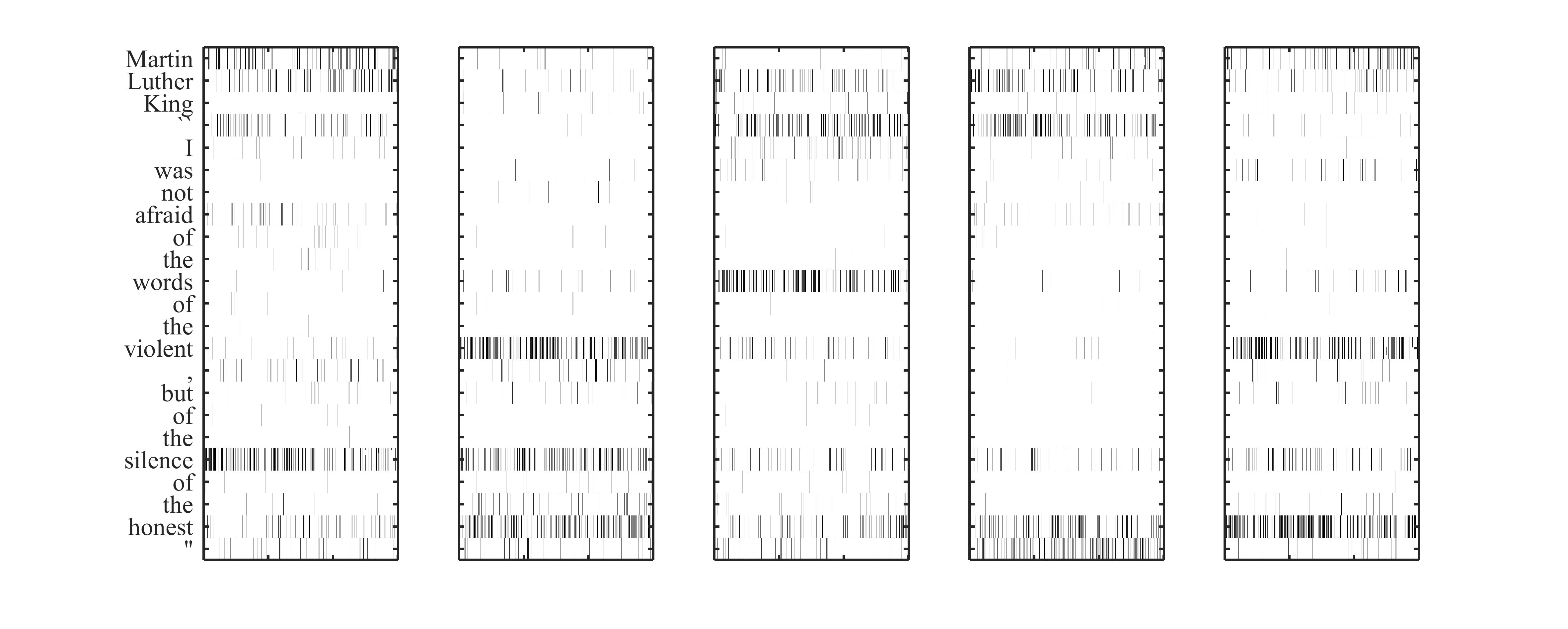}
  \caption{Age dataset with penalization}
\end{subfigure}% 
\\
\begin{subfigure}{1\textwidth}
  \centering
  \includegraphics[width=1\linewidth]{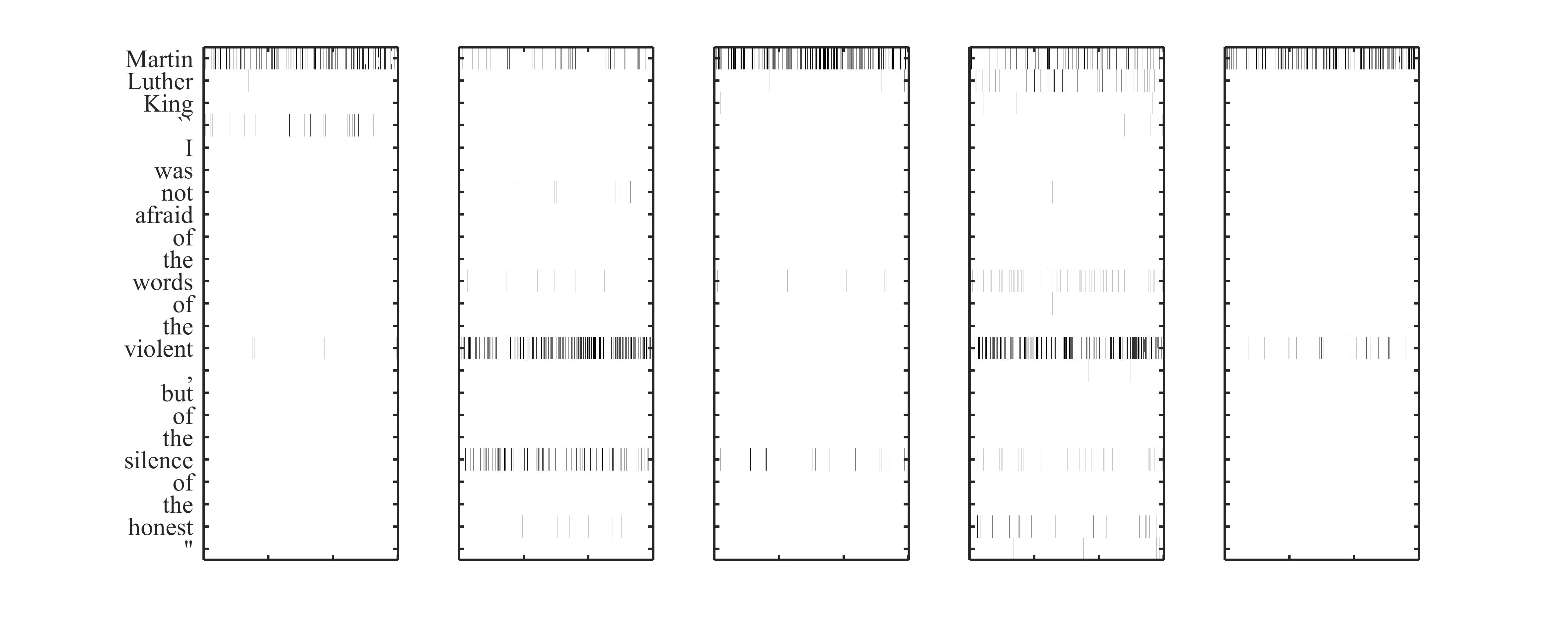}
  \caption{Age dataset without penalization}
\end{subfigure}
	\caption{Visualization of vectorial multi-head attention. The vertical and horizontal axes indicate the source word tokens and the 600 dimensions of the attention ${\mat A}^i$ for different heads. }
	\label{fig:view}
\end{figure}

To verify whether the penalization term $P$ discourages the redundancy in the sentence embedding, we visualize the vectorial multi-head attention according. We compare two models with the same hyper-parameters except that one is with penalization on attention matrices and the other without penalization. We pick a sentence from the development set of the Age data: \textit{Martin Luther King ``I was not afraid of the words of the violent, but of the silence of the honest'' }, with the gold label being the category of 65+. We plot all 5 heads of attention matrices as in Figure~\ref{fig:view}. From the figure we can tell that the model trained without the penalization term has much more redundancy between different heads of attention (Figure 3b), resulting in putting significant focus on the word ``Martin'' in the 1st, 3rd and 5th head, and on the word ``violent'' in the 2nd and 4th head. However in Figure 3a, the model with penalization shows much more variation between different heads.  

\begin{table*}[t!]
\centering
\scalebox{1}{
\begin{tabular}{|l|c|c|c|c|c|}
\hline
\textbf{Model} & \textbf{SNLI} & \multicolumn{2}{c|}{\textbf{MultiNLI}} & \textbf{Yelp}& \textbf{Age}   \\
 &  & \textbf{IN} & \textbf{Cross} &  &  \\
\hline
w/ Penalization on Parameter Matrices & \textbf{86.6}  & \textbf{73.8} & \textbf{74.0} & \textbf{66.55} & 82.45\\
w/ Penalization on Attention Matrices &  86.2 & 73.6  & 73.8  & 66.15 & \textbf{82.63}      \\
w/ Penalization on Sentence Embeddings & 86.1  & 73.5 & 73.6 & 65.75  & 82.15 \\
w/o Penalization & 86.4  & 73.7 & 73.4 & 65.25  & 82.18\\ 
\hline
\end{tabular}
}
\caption{Performance with/without the penalization term. The penalization weight is selected from [1,1e-1,1e-2,1e-3,1e-4] on the development sets.}
\label{tab:penalization}
\end{table*}

\section{Conclusions}

In this paper, we propose a generalized pooling method for sentence embedding through vector-based multi-head attention, which includes the widely used max pooling, mean pooling, and scalar self-attention as its special cases. Specifically the proposed model aims to use vectors to enrich the expressiveness of attention mechanism and leverage proper penalty terms to reduce redundancy in multi-head attention. We evaluate the proposed approach on three different tasks: natural language inference, author profiling, and sentiment classification. The experiments show that the proposed model achieves significant improvement over strong sentence-encoding-based methods, resulting in state-of-the-art performances on four datasets. The proposed approach can be easily implemented for more problems than we discuss in this paper. 

Our future work includes exploring more effective MLP to use the structures of multi-head vectors, inspired by the idea from~\newcite{DBLP:journals/corr/LinFSYXZB17}. Leveraging structure information from syntactic and semantic parses
is another direction interesting to us.

\section*{Acknowledgments}
This work was partially funded by the National Natural Science Foundation of China (Grant No. U1636201) and the Key Science and Technology Project of Anhui Province (Grant No. 17030901005).
% The acknowledgments should go immediately before the references. Do not include this section when submitting your paper for review.

\clearpage
\bibliographystyle{acl}
\bibliography{coling2018}

\end{document}